%% file: main.tex
\documentclass[10pt,twocolumn,letterpaper]{article}

\usepackage{multirow}
\usepackage{adjustbox}
\usepackage[table]{xcolor}
\definecolor{lightorange}{RGB}{255, 230, 200}

\usepackage[pagenumbers]{cvpr} %

\input{preamble}

\newcommand{\dataname}[0]{WMGStereo-150k}
\newcommand{\parbf}[1]{\par \noindent \textbf{#1}.}

\definecolor{cvprblue}{rgb}{0.21,0.49,0.74}
\usepackage[pagebackref,breaklinks,colorlinks,allcolors=cvprblue]{hyperref}

\def\paperID{17691} %
\def\confName{CVPR}
\def\confYear{2026}

\title{What Makes Good Synthetic Training Data for Zero-Shot Stereo Matching?}

\author{David Yan\\
Princeton University\\
{\tt\small yan.david@princeton.edu}
\and
Alexander Raistrick\\
Princeton University\\
{\tt\small araistrick@princeton.edu}
\and
Jia Deng\\
Princeton University\\
{\tt\small jiadeng@princeton.edu}
}

\begin{document}
\maketitle

\begin{abstract}

Synthetic datasets are a crucial ingredient for training stereo matching networks, but the question of what makes a stereo dataset effective remains underexplored. We investigate the design space of synthetic datasets by varying the parameters of a procedural dataset generator, and report the effects on zero-shot stereo matching performance using standard benchmarks. We validate our findings by collecting the best settings and creating a large-scale dataset. Training only on this dataset achieves better performance than training on a mixture of widely used datasets, and is competitive with training on the FoundationStereo dataset, with the additional benefit of open-source generation code and an accompanying parameter analysis to enable further research.  We open-source our system at \href{https://github.com/princeton-vl/InfinigenStereo}{this URL} to enable further research on procedural stereo datasets.

\end{abstract}

\section{Introduction}
\label{sec:intro}

Synthetic data rendered by computer graphics is widely used for training models for stereo matching. Stereo matching uses a pair of RGB images to estimate per-pixel disparity, which can in turn be used to calculate scene depth. Synthetic datasets provide high-quality ground truth disparity calculated using depth from the rendering pipeline. State-of-the-art methods in stereo matching \cite{mayer2016large, chang2018pyramid, yin2019hierarchical,croco_v2, xu2023iterativegeometryencodingvolume, guan2024neural, zeng2024temporally, lipson2021raft, li2022practicalstereomatchingcascaded, wang2024selective, zhao2023high, xu2024igev++} all train on synthetic datasets \cite{mayer2016large,  gaidon2016virtual, tremblay2018falling, yang2019hierarchical, cabon2020virtual, wang2020tartanairdatasetpushlimits}.  Therefore, developing new synthetic datasets has been a significant component of recent work on stereo matching \cite{li2022practicalstereomatchingcascaded, wen2025foundationstereozeroshotstereomatching}.

Designing effective synthetic datasets remains a significant challenge. Existing synthetic stereo datasets vary significantly in their designs and realism, from random flying objects scattered in a scene \cite{mayer2016large, li2022practicalstereomatchingcascaded}, to realistic domain-specific simulators \cite{gaidon2016virtual, cabon2020virtual}. However, assessing \textit{what matters in dataset design} is difficult because new datasets typically change many different factors at once. If an existing dataset is particularly helpful, was it because of the scene realism, object diversity, materials, or camera placement? The best choices are not obvious - more extreme arrangements could either improve a model's performance by making it more robust to a wider distribution, or could harm performance by diverging from real-world scenes in downstream tasks.

In this work, we systematically explore the design space of synthetic data for stereo matching using procedural generation. Procedural generation creates synthetic data using randomized computer graphics algorithms, which means the entire data generation process can be controlled and customized to investigate downstream performance. 

We construct a procedural generator for stereo datasets by combining object- and scene-level generators from Infinigen \cite{infinigen2023infinite, infinigen2024indoors} with new features designed to support large-scale stereo dataset generation. To ensure good coverage of common stereo dataset designs, we add new placement generators for floating objects and random lighting. We also add tools to avoid problematic objects and materials, such as one that removes glass from subparts of objects. By varying the procedural parameters of this generator, we can analyze several common features of stereo datasets, including floating object density, background objects, object category, object materials, and augmented lighting, as well as the parameters of these features, e.g. the density of floating object placement, or which materials are used for augmentation.

\begin{figure*}
    \centering
    \includegraphics[width=1\textwidth]{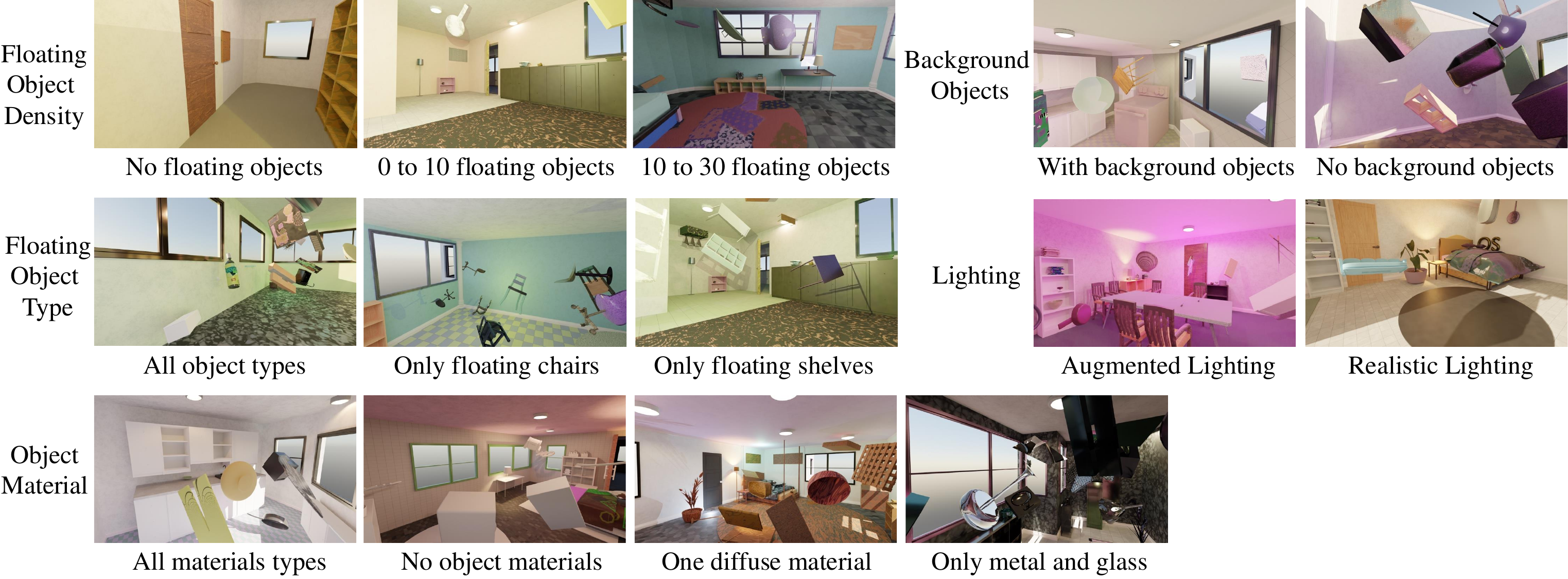}
    \caption{\textbf{Visualization of synthetic dataset design choices}. For each studied parameter, we show a single illustrative example demonstrating the effect of changing dataset generation parameters such as object density, background realism, object types, lighting, and materials. Each example is sampled from a dataset evaluated in Tab. \ref{tab:parameter_analysis_quantitative}.}
    \label{fig:ablation_examples}
\vspace{-5pt}
\end{figure*}

We investigate which dataset designs lead to strong benchmark performance in zero-shot, that is, \textit{without} any fine-tuning. For each parameter setting, we generate a stereo matching dataset, then train and evaluate a RAFT-Stereo model to determine the downstream effect on zero-shot performance. We evaluate in the zero-shot setting, similar to FoundationStereo \cite{wen2025foundationstereozeroshotstereomatching} and StereoAnywhere \cite{bartolomei2025stereoanywhererobustzeroshot}, as it represents the typical off-the-shelf usage of a model. We aim to improve zero-shot performance on existing models by training on diverse synthetic data.

Our study yields a variety of interesting findings. Of all the designs tested, we found the most effective stereo data comes from the \textit{combination} of realistic indoor scenes with added floating objects. Removing realistically arranged background furniture (leaving floating objects in empty rooms) or removing the entire background (leaving just floating objects) harms zero-shot generalization. These results show that certain aspects of scene realism are helpful for zero-shot stereo, while also suggesting that scenes designed to be fully realistic lack sufficient geometry diversity to be data-efficient. We also find that existing networks struggle to learn reflective and transparent materials without large performance drops on diffuse regions, suggesting that co-design of architecture and data might be necessary for robust non-Lambertian stereo. Finally, we find that varying camera baseline greatly improves generalization, and that ray-tracing quality can be traded off for more data.

We use the best discovered settings to create \dataname{}, a new training dataset for stereo matching. Training on only this dataset produces better zero-shot results than training on SceneFlow \cite{mayer2016large}, CREStereo \cite{li2022practicalstereomatchingcascaded}, TartanAir \cite{wang2020tartanairdatasetpushlimits}, or IRS \cite{wang2021irs}. Moreover, training DLNR \cite{zhao2023high} on \dataname{} outperforms training on the \textit{combination} of these prior datasets by 28\% on Middlebury \cite{Scharstein2014MiddleburyStereo} and by 25\% on the Booster \cite{zamaramirez2024booster} evaluation sets (Tab. \ref{tab:stereo_results}). We achieve competitive performance with the state-of-the-art dataset, FoundationStereo \cite{wen2025foundationstereozeroshotstereomatching} (Tab. \ref{tab:dataset_results}). Our dataset is also highly sample efficient: training on just 500 random examples from our dataset achieves lower error on Middlebury than 100,000 CREStereo examples (Fig. \ref{fig:scaling}).

In summary, we make three main contributions. First, we provide a thorough analysis of the effects of common stereo dataset designs and possible parameter settings. Second, we provide \dataname{}, a new large-scale dataset incorporating these findings. Finally, and uniquely among stereo datasets, our work provides procedural generation code, which we hope will be useful to generate additional data or enable further research to improve the usefulness of synthetic datasets for stereo matching.

\section{Related Work}

\begin{figure*}
    \centering
    \includegraphics[width=\textwidth,height=0.2\textheight]{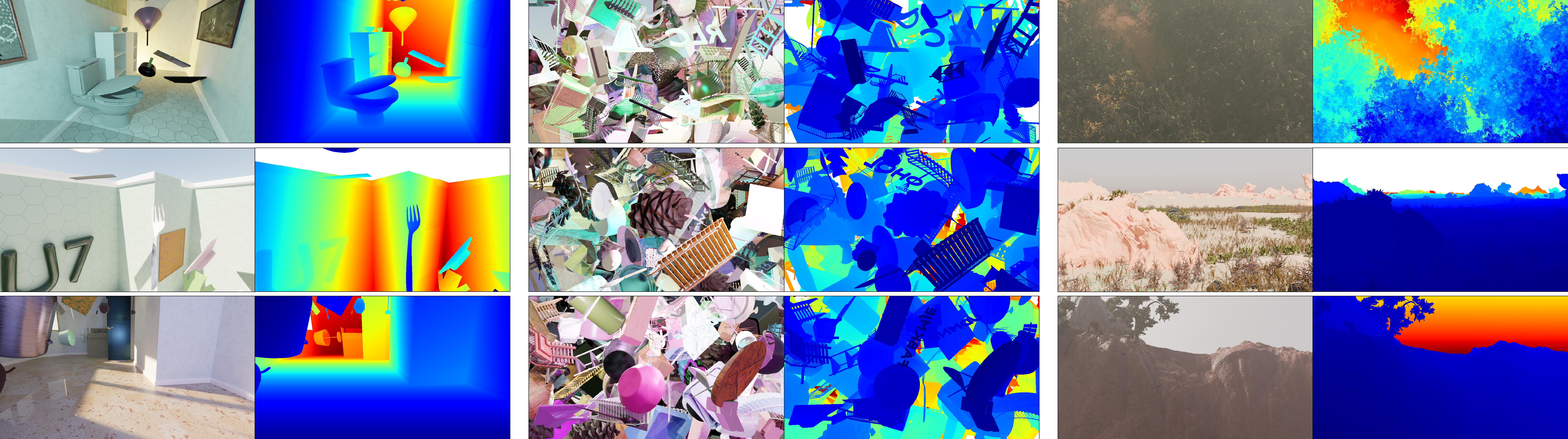}
    \caption{\textbf{\dataname{} Dataset}. From left to right, we show random, non-cherrypicked samples from our \textit{Indoors with Floating Objects}, \textit{Dense Floating Objects}, and \textit{Nature} scene types. Refer to the appendix for additional samples.} 
    \label{fig:dataset_examples}
\vspace{-8pt}
\end{figure*}

\textbf{Synthetic Datasets for Stereo Matching} Procedural datasets such as SceneFlow \cite{mayer2016large} typically use randomly placed flying objects with randomized backgrounds to achieve high diversity. This procedural recipe for object placement remains highly effective, with procedural datasets such as CREStereo \cite{li2022practicalstereomatchingcascaded} and FallingThings \cite{tremblay2018falling} building upon the flying/floating object setup with additions such as photorealistic rendering and more diverse materials.

Non-procedural synthetic datasets \cite{butler2012naturalistic, Mehl2023} such as IRS \cite{wang2021irs}, TartanAir  \cite{wang2020tartanairdatasetpushlimits}, TartanGround \cite{patel2025tartangroundlargescaledatasetground}, XR-Stereo \cite{cheng2023stereomatchingtime100}, DISC \cite{DBLP:conf/iros/JeonILCHK19}, UnrealStereo4k \cite{DBLP:journals/corr/abs-2104-03866}, and DynamicReplica \cite{Karaev_2023_CVPR} typically render frames from realistic scenes handcrafted by 3D artists. Simulators such as HS-VS \cite{yang2019hierarchical} and VirtualKITTI \cite{gaidon2016virtual} \cite{cabon2020virtual} primarily target driving scenarios.

Although these datasets have been widely adopted and have demonstrated success in state-of-the-art stereo networks, they do not report ablations over dataset generation parameters, and therefore provide limited insight into what actually makes these datasets so effective. Moreover, we cannot directly study generation parameters in existing datasets (e.g. re-rendering TartanAir with different materials) because they do not open-source their generation code or assets. We use the open-source Infinigen system to reproduce and analyze existing design choices. We perform a detailed analysis of what parameters, such as object placement and materials, matter most for downstream performance. 

FoundationStereo \cite{wen2025foundationstereozeroshotstereomatching} achieves state-of-the-art zero-shot performance by introducing a novel architecture alongside the FoundationStereo dataset (FSD). FSD simultaneously introduces many new dataset features, such as varied camera parameters, flying objects, randomized lighting, realistic background scenes, and physics-based simulation. Our parameter study disentangles and identifies which of these design choices are most critical for downstream performance, informing future dataset design. FSD is also a static dataset, while our dataset generation code is open-source, allowing our dataset to be expanded and tailored for target use cases. A user targeting non-Lambertian stereo could easily regenerate our data with only glass materials, while FSD provides limited coverage of transparent objects. Our dataset generator therefore uniquely enables the co-design of data with new tasks and model architectures.

\textbf{Procedural Generation}
Although there exist several open platforms for procedural data generation, there has been limited use of these platforms to produce large-scale stereo datasets. Kubric \cite{greff2021kubric} generates scattered or falling objects in a 3D scene, while ProcTHOR \cite{procthor} generates realistic indoor environments for embodied agent training. However, neither pipeline has demonstrated use in stereo matching. Infinigen Nature \cite{infinigen2023infinite} and Infinigen Indoors \cite{infinigen2024indoors} use Blender \cite{blender} to procedurally generate realistic indoor and natural scenes. Raistrick et al. \cite{infinigen2023infinite} created a stereo dataset using Infinigen Nature, but it did not achieve competitive performance with SceneFlow. The Infinigen-SV \cite{jing2024matchstereovideos} dataset only uses Infinigen Nature and focuses on augmenting SceneFlow for video-consistent stereo networks.

Our work significantly expands on the object and scene generators from Infinigen \cite{infinigen2023infinite, infinigen2024indoors}. We develop a procedural placement algorithm for floating object and lighting placement integrated with Infinigen scene and object generators. Our algorithm supports common features in stereo datasets and enables us to exercise fine-grained control over objects, materials, lighting, and cameras. For instance, we can remove different classes of objects from generation and even remove specific materials from parts of an individual objects. Our dataset is a novel demonstration of general-purpose scene generators for stereo data.

\textbf{Analysis of Synthetic Datasets} Mayer et al. \cite{Mayer_2018} perform a detailed study of the procedural generation parameters of synthetic training data for optical flow and disparity prediction. This study concludes that ``realism is overrated" for effective synthetic data, motivating the further development of successful flow datasets such as AutoFlow \cite{sun2021autoflow, huang2023self}. However, this study primarily studies generation parameters for FlyingChairs-style datasets  \cite{DFIB15, ISKB18}, which consist only of warped 2D images with no depth or disparity. Moreover, their study of disparity prediction focuses on augmentations simulating camera noise and blurring. As such, there is limited study of critical generation parameters, such as object placement and realism, for modern stereo datasets rendered from 3D scenes. More recently, StereoCarla \cite{guo2025stereocarlahighfidelitydrivingdataset} studied the effects of camera placement and weather effects for driving simulator datasets.

In contrast to Mayer et al., our study
focuses on stereo depth prediction, uses more realistic 3D
procedural generators, and applies directly to modern stereo
networks. We find that a mix of placement realism and random floating objects is an effective recipe for stereo data. While this result supports Mayer et al.'s conclusion that diversity from randomly scattered objects is important, it also suggests that scene realism can help significantly with zero-shot generalization. We also study both high-level design choices such as floating object placement and low-level parameters such as materials.

\begin{table*}
    \centering
    \scriptsize
    \setlength{\tabcolsep}{3pt}
    \renewcommand{\arraystretch}{1.2}
    \begin{tabular}{llccccccc}
        \toprule
        \cmidrule(lr){3-9}
        Experiment & Method & Middlebury 2014 (F) & Middlebury 2014 (H)  & Middlebury 2021 & ETH3D  & KITTI-12& KITTI-15 & Booster (Q) \\
        \midrule
        \multirow{3}{*}{Floating Object Density} 
        & No floating objects & 18.44 & 12.52 & 16.31 & 4.47 & 4.42 & 6.19 & 16.40 \\
        & 0 to 10 floating objects & 11.42 & 7.78 & 11.30 & \textbf{3.62} & 4.44 & 6.09 & 12.21 \\
        & \textbf{10 to 30 floating objects} & \textbf{9.19} & \textbf{6.60 }& \textbf{10.28} & 3.92 & \textbf{4.05 }& \textbf{5.11 }& \textbf{10.60 }\\ 
        \midrule
        \multirow{2}{*}{Background Objects} 
        & No background objects & 10.42 & 8.35 & 12.01 & 4.39 & 4.20 & 6.28 & 12.72 \\
        & \textbf{With background objects} & \textbf{9.19} & \textbf{6.60 }& \textbf{10.28} & \textbf{3.92} & \textbf{4.05} & \textbf{5.11} & \textbf{10.60} \\
        \midrule
        \multirow{3}{*}{Floating Object Type} 
        & Floating chairs & 9.24 & \textbf{5.29} & 9.81 & 3.64 & 4.90 & 7.02 & 11.22 \\
        & Floating shelves & 10.28 & 7.13 & 10.24 & 3.51 & 4.32 & 6.06 & 11.63 \\
        & Floating bushes & 9.20 & 6.04 & \textbf{9.53} & \textbf{3.13} & \textbf{3.86} & 5.42 & 12.19\\
        & \textbf{All generators used} & \textbf{9.19} & 6.60 & 10.28 & 3.92 & 4.05 & \textbf{5.11 }& \textbf{10.60} \\
        \midrule
        \multirow{4}{*}{Object Material} 
        & No materials & 11.42 & 9.02 & 10.65 & 3.48 & 4.34 & 6.07 & 14.07 \\
        & One diffuse material & 10.09 & 7.21 & \textbf{9.65} &\textbf{2.77} & \textbf{3.76} & 5.41 & 12.73 \\
        & Only metal and glass & 11.28 & 8.37 & 11.85 & 4.95 & 4.06 & \textbf{4.97} & \textbf{9.80} \\
        & \textbf{All materials used} &\textbf{9.19} & \textbf{6.60} & 10.28 & 3.92 & 4.05 & 5.11 & 10.60 \\
        \midrule
        \multirow{3}{*}{Stereo Baseline} 
        & Uniform[0.04, 0.1] & 32.47 & 9.60 & 22.18 & \textbf{2.89 }& 5.13 & 6.64 & 17.03 \\
        & Uniform[0.2, 0.3] & 9.75 & 7.01 & 10.50 & 14.05 & \textbf{3.94} & 5.37 & \textbf{8.96} \\
         & \textbf{Uniform[0.04, 0.4]}& \textbf{9.19} & \textbf{6.60} & \textbf{10.28} & 3.92 & 4.05 & \textbf{5.11 }& 10.60 \\
        \midrule
        \multirow{2}{*}{Lighting} 
        & Realistic Lighting & 9.62 & 6.91 & \textbf{10.06} & \textbf{3.81} & \textbf{3.95 }& 5.45 & \textbf{10.45} \\
        & \textbf{Augmented Lighting} & \textbf{9.19 }& \textbf{6.60}& 10.28 & 3.92 & 4.05 & \textbf{5.11 }& 10.60 \\

\bottomrule
    \end{tabular}
    \caption{\textbf{Results of generation parameter study}. For each variation, we generate a new dataset of 5,000 stereo pairs and use it to train RAFT-Stereo \cite{lipson2021raft}. We report the zero-shot performance on a variety of standard benchmarks. We bold the best-performing models and the settings chosen for our final generator.}

    \label{tab:parameter_analysis_quantitative}
\vspace{-7pt}
\end{table*}

\section{Analysis of Dataset Parameters}
\label{sec:analysis}

\subsection{Initial procedural generator}

We construct our system on top of the Infinigen and Blender Python APIs. Infinigen provides high-level APIs to generate objects (e.g. chairs, plants, staircases) and scenes (indoor rooms, nature, or blank sky backgrounds). We directly use the Blender API to implement multiple new scene arrangement generators and other modifications that are specifically tailored to create effective stereo datasets. 

Specifically, we create a floating object placement interface that calls the Infinigen API to generate objects and then arranges them throughout a given scene type. Our floating object placement has options to place objects in the view of a given camera through raycasting or to place objects within a specified bounding box. We also provide an option to allow floating objects to intersect existing scene geometry.

Our modifications to the Infinigen system and our floating object interface enable the system to generate three distinct scene types: indoor floating objects, dense floating objects, and nature. For the indoor floating objects scene type, we generate rooms from the Infinigen Indoors system and use our interface to generate objects at randomly sampled locations in the room. For the dense floating object scene type, we spawn a camera rig in an empty sky scene and use our interface to place objects within view of the camera. Nature scenes are generated from Infinigen Nature. 

\subsection{Parameter Study}

We perform a systematic study to determine the effect of procedural generation parameters on zero-shot stereo matching performance. For a variety of parameters, we directly evaluate their effect on zero-shot performance. In each such case, we generate 5000 stereo pairs using the indoor floating objects scene type. We show a single illustrative example of each dataset in Fig. \ref{fig:ablation_examples}. We train RAFT-Stereo \cite{lipson2021raft} on each of these datasets from a random initialization for 75k steps using standard hyperparameters and augmentation. Finally, we evaluate the resulting models on the Middlebury \cite{Scharstein2014MiddleburyStereo}, ETH3D \cite{schoeps2017cvpr}, KITTI \cite{geiger2012we, menze2015object}, and Booster \cite{zamaramirez2024booster} datasets and report results in Tab. \ref{tab:parameter_analysis_quantitative}. This training setup uses less data and a shorter training schedule to enable fast experimentation. In practice, this reduced scale is sufficiently representative of the full-scale training runs.

We evaluate zero-shot performance on the Middlebury 2014 evaluation set, the Middlebury 2021 dataset, the ETH3D train set, the KITTI 2012 and KITTI 2015 train sets, and the Booster train set. The px error is defined as the percentage of pixels with end-point-error
exceeding a specified threshold. The thresholds are chosen using standard ranking metrics: 2px non-occluded error for Middlebury 2014, 2px error for Middlebury 2021, 1px error on ETH3D, 3px error for KITTI 2012 and KITTI 2015, and 2px error on Booster. We evaluate on full (F) and half (H) resolution on Middlebury and on quarter (Q) resolution on Booster. 

\parbf{Floating Object Density} Floating objects (or flying objects) are a common feature of stereo datasets. However, it is plausible that realistically arranged objects such as those produced by Infinigen Indoors would be better aligned with the distribution of real-world scenes. First, we validate whether adding floating objects to scenes that already contain realistic room layouts provides any benefit. We compare scenes with either no floating objects or 0 to 10 floating objects, and confirm that this improves Middlebury 2014 (H) 2px error (12.52 to 7.78). Next, we compare this to a setting with far greater density of floating objects (uniformly sampled from 10 to 30 per scene) and again find a 2px error improvement (7.78 to 6.60). This confirms that floating objects are extremely helpful for zero-shot performance, despite resulting in less realistic scenes. Following this result, we place 200 objects in our dense floating object scenes to maximize object density. 

\parbf{Background Objects} Next, we investigate whether the furniture and other background objects (e.g. shelves, tables) which appear in indoor scenes are useful. Such arranged objects are not present in other procedural datasets like FlyingThings3D, and previous studies indicate that realism, such as these realistically placed objects, is not important \cite{Mayer_2018}. However, when evaluating scenes with and without background objects, we find that including background objects \textit{does} help, improving performance on all benchmarks (e.g. 8.35 to 6.60 2px error on Middlebury 2014 (H)).

\begin{table*}
\centering
\begin{tabular}{lcccccc}
\toprule
Model & Middlebury 2014 (H) & Middlebury 2021 & ETH3D & \multicolumn{2}{c}{KITTI} & Booster (Q) \\
      &                    &                 &       & 2012  & 2015  &  \\
\midrule
PSMNet \cite{chang2018pyramid}                   & 13.80  & 23.67 & 19.75 & 6.73  & 6.78  & 34.47 \\
RAFT-Stereo \cite{lipson2021raft}                & 8.66  & 10.28 & 2.6   & 4.35  & 5.67  & 17.37 \\
IGEV \cite{xu2023iterativegeometryencodingvolume}& 11.81 & 20.43 & 43.05 & 7.62  & 7.81  & 23.38 \\
GMStereo \cite{xu2023unifying}                   & 10.98 & 25.43 & 6.22  & 5.68  & 5.72  & 32.44 \\
DLNR \cite{zhao2023high}                         & 6.20   & 8.44  & 23.01 & 9.08  & 16.05 & 18.15 \\
Selective-IGEV  \cite{wang2024selective}         & 7.30   & 8.97  & 6.07  & 5.64  & 6.05  & 17.58 \\
NMRF  \cite{guan2024neural}                      & 10.87 & 23.36 & 4.34  & 4.62  & 5.24  & 27.08 \\
IGEV++ \cite{xu2024igev++}                       & 7.75  & 9.58  & 4.67  & 6.23  & 6.40  & 17.22 \\
StereoAnywhere \cite{bartolomei2025stereoanywhererobustzeroshot} & 4.75 & 5.71 & 1.43 & 3.52 & 3.79 & 9.01 \\
FoundationStereo \cite{wen2025foundationstereozeroshotstereomatching} & \textbf{1.10} & \textbf{4.17} & \textbf{0.50} & \textbf{2.30} & \textbf{2.80} & \textbf{4.16} \\
\midrule
RAFT-Mixed                                & 5.50 & 8.97 & \textbf{2.58} & 3.64 & 4.95 & 11.46  \\
\rowcolor{lightorange}
RAFT-\dataname{}                          & \textbf{4.48}  & \textbf{8.17}  & 2.93  & \textbf{3.25}  & \textbf{4.25}  & \textbf{9.17}  \\
\midrule
Selective-IGEV-Mixed                      & 5.24 & 8.24 & \textbf{2.37} & 3.97 & 5.31 & 11.00 \\
\rowcolor{lightorange}
Selective-IGEV-\dataname{}                & \textbf{3.61}  & \textbf{7.62} & 2.47 & \textbf{3.26} & \textbf{4.55} & \textbf{8.84}
\\
\midrule
DLNR-Mixed                                & 5.21  & 9.30   & \textbf{2.50}   & 3.68  & 4.95  & 12.17 \\
\rowcolor{lightorange}
DLNR-\dataname{}                          & \textbf{3.76}  & \textbf{6.72}  & \textbf{2.50}  & \textbf{3.30}  & \textbf{4.54}   & \textbf{9.09} \\
\bottomrule
\end{tabular}
\caption{\textbf{Zero-shot stereo results} Models trained on \dataname{} outperform many prior methods (top section) on zero-shot stereo matching for standard benchmarks. Mixed is a combined synthetic dataset of SceneFlow, CREStereo, TartanAir, and IRS (600k total pairs). Training on our dataset yields improvements for many stereo models, demonstrating that our findings generalize across architectures. Although our models perform worse than  FoundationStereo, which is significantly larger than our retrained models, we achieve competitive performance with StereoAnywhere, a recent model focused on zero-shot stereo with similar computational cost to our tested architectures.}
\vspace{-9pt}
\label{tab:stereo_results}
\end{table*}

\parbf{Object Type} We investigate to what extent the diversity of objects is useful: how helpful is sampling from all available Infinigen object generators, as opposed to using only a single class of objects? We find that using only chairs results in small improvements on some benchmarks (Middlebury 2014 and 2021). Fixing the object type to shelves results in worse performance on all benchmarks except ETH3D. Meanwhile, using only the bush object type leads to better performance on KITTI-12 and ETH3D. These results likely reflect the biases of individual benchmarks toward certain object types. For instance, one can improve performance on indoor benchmarks (6.60 to 5.29 2px error on Middlebury 2014 (H)) by using only chairs, but at the cost of significantly worse performance on driving scenarios (5.11 to 7.02 3px error on KITTI-15). Meanwhile, using only bushes helps on benchmarks like KITTI-12 and ETH3D (3.92 to 3.13 1px error), which have more nature coverage but hurts performance on Booster (10.60 to 12.19 2px error), which focuses on specular and indoor scenes. As such, we choose to use all generators because it has the most robust performance across many benchmarks.

\begin{figure}
    \centering
    \includegraphics[width=1\linewidth]{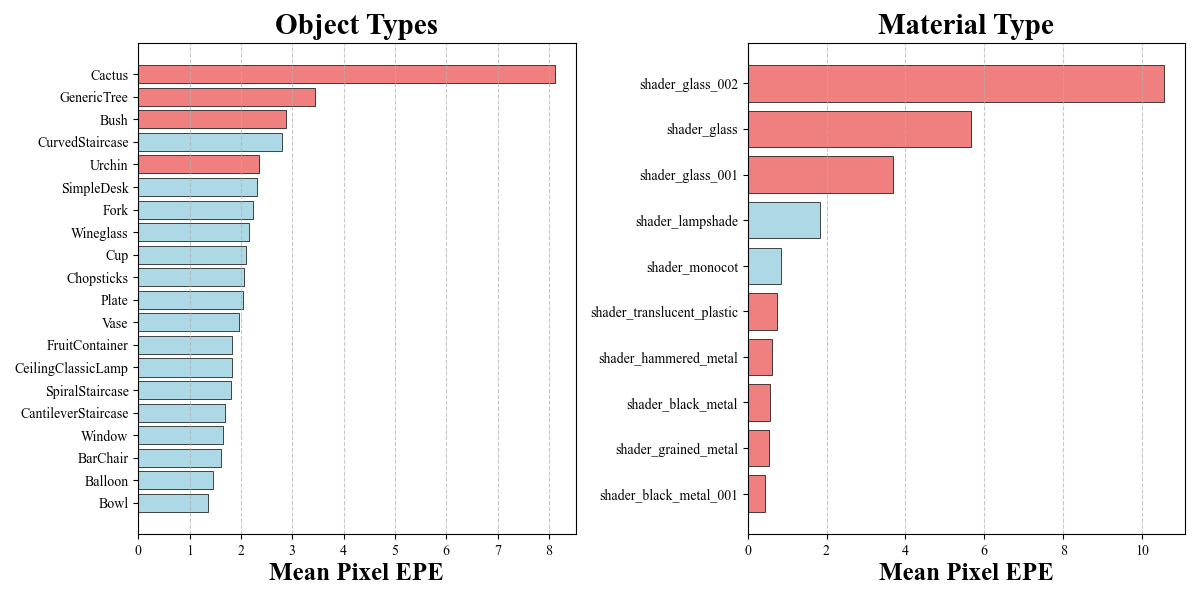}
    \caption{\textbf{End-point-error averaged by object and material}. We show average error only for the materials that make up at least 0.1\% of pixels. Assets marked in red were removed after manual inspection because they introduce ambiguous high-error cases, such as completely reflective surfaces, or imperceptible holes.}
    \label{fig:epe_by_obj_material}
\vspace{-5pt}

\end{figure}

We also investigated objects at the level of individual object types and per-pixel errors. Specifically, we test a trained model on 1000 additional images rendered from the same procedural parameters, and we use object segmentation masks to compute average end-point-error aggregated across each object type. We plot the objects with the highest remaining error in Fig. \ref{fig:epe_by_obj_material}. We manually inspected error maps for the highest error objects, and found objects such as cacti and urchins, which have very thin needle structures, and racks, which have very small holes. We remove these objects from our final system due to their extreme difficulty.

\parbf{Object Materials} We investigate the impact of various categories of materials. We applied changes to all objects except the walls to prevent very ill-posed cases such as transparent or mirror-like walls. When only glass and metallic materials are used, the model achieves the best performance on KITTI-15 and Booster, likely because these benchmarks feature many transparent or reflective surfaces. However, it has significantly worse performance on Middlebury and ETH3D. When using just a single diffuse material such as wood, the model obtains the best performance on Middlebury 2021, ETH3D, and KITTI-12, but does poorly on Booster, likely because of the lack of glass/metal coverage in the training data. When using no materials at all, the model performs worse across all benchmarks except ETH3D. Using a diverse array of materials results in the most robust performance across all benchmarks.

Similarly to object types, we also group per-pixel EPE by material and plot the materials with the highest error in Fig. \ref{fig:epe_by_obj_material}. We limit this analysis to materials which comprise at least 0.1\% of pixels, to prevent very rare materials (e.g. creature's tongue / eyeball) from dominating the statistics due to low sample count. Eight of the ten highest errors are troublesome versions of glass and metal, which are often completely transparent or completely reflective. We remove these specific materials from our system, because we find current stereo networks struggle to handle ill-posed surfaces without degrading performance on diffuse regions. However, we retain many other non-Lambertian materials which are only moderately transparent/reflective. We take particular care for external windows - replacing the glass material for these with an opaque surface would ruin lighting for the rest of the scene, so we instead automatically delete the geometry to eliminate it from the ground truth.

\parbf{Lighting Augmentation} We perform a suite of different lighting augmentations. We randomly place 0 to 5 point lights in the scene, with a total power sampled uniformly from 250 to 1250. We also randomly remove the ceiling with probability 0.3 to allow occasional natural sky lighting, and randomly dim the ceiling lights and alter their colors with probability 0.2 to cover darker scenes. 

We found that lighting augmentation had little effect on benchmark performance, with some improvements on Middlebury 2014 and KITTI-15, and slightly worse performance on ETH3D, Middlebury 2021, and KITTI-12. It is plausible that the lack of difference is caused by insufficient lighting variation in the benchmarks. We opt to include augmented lighting in our dataset to better account for diverse, in-the-wild lighting variations.

\parbf{Stereo baseline randomization} We find that having a wide range of camera baseline values is very significant for robust generalization. When trained only on small baselines uniformly sampled between 0.04 and 0.1 m, the model performs significantly worse across all benchmarks except ETH3D. When trained only on large baselines sampled between 0.2 and 0.3 m, the model's performance improves on certain large disparity benchmarks such as Booster, but fails on small disparity benchmarks such as ETH3D. We choose to sample the baseline value uniformly from 0.04 to 0.4 m, which had the best performance.

\begin{table}[ht]
\centering
\scriptsize
  \begin{tabular}{lcccccc}
  \toprule
  Data Mix & \multicolumn{2}{c}{Middlebury 2014} & Middlebury  & ETH3D & \multicolumn{2}{c}{KITTI}  \\
                   & (F) & (H) & 2021 &   & 12 & 15   \\
  \midrule
  Indoor Floating                         & 9.19  & 6.60   & 10.28 & 3.92  & 4.05  & \textbf{5.11}   \\
  Dense Floating                         & 15.25 & 9.90   & 12.25 & 3.40   & 7.22  & 9.13  \\
  Nature                         & 18.61 & 12.27 & 16.92 & 4.23  & 6.24  & 7.74  \\
    \textbf{Mixed Dataset} & \textbf{7.81 } & \textbf{6.04}  & \textbf{9.29}  &\textbf{ 2.51 } & \textbf{3.54}  & 5.12   \\
  \midrule
  10 - 80 - 10                   & 10.52 & 6.09 & 10.05 & 2.66  & 3.79  & \textbf{4.86}   \\
  80 - 10 - 10                   & 8.13  & \textbf{5.90 }  & \textbf{9.21 } & 3.01  & 3.81  & 5.31  \\
  10 - 10 - 80                   & 9.70   & 7.14  & 10.63 & 2.79  & 3.83  & 5.50  \\
    \textbf{33 - 33  - 33  }                & \textbf{7.81}  & 6.04  & 9.29  & \textbf{2.51 } & \textbf{3.54}  & 5.12   \\
  \bottomrule
  \end{tabular}

\caption{\textbf{Performance of different data distributions}. Top half: individual scene types. Bottom half: different weightings of scene types. Mixed denotes a dataset of equal parts indoors, dense floating. Each label denotes \% Indoor - \% Dense Floating - \%Nature data. A mixed dataset of all dataset types is the most effective, while the indoor floating type is the best individual data type.}

\label{tab:data_distribution}
\end{table}

\begin{figure*}
    \centering
    \includegraphics[width=1\linewidth]{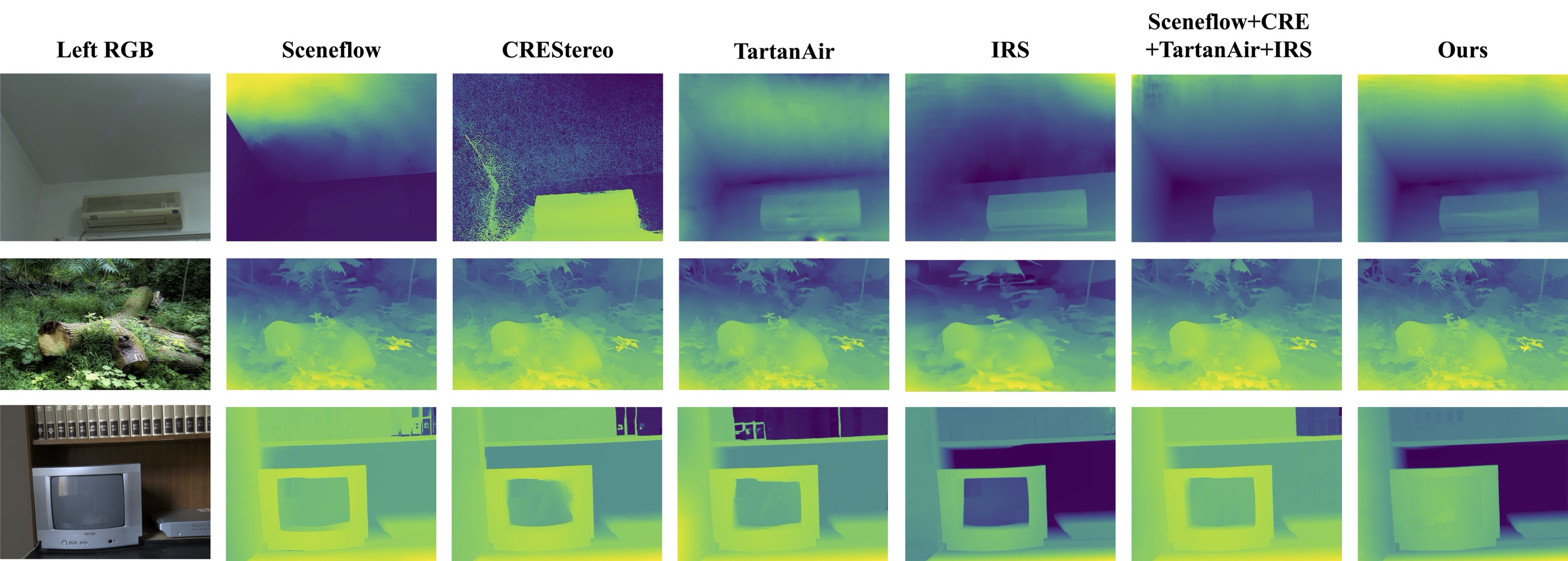}
    \caption{\textbf{Qualitative comparison of in-the-wild predictions}. We train DLNR on \dataname{} and existing synthetic datasets. Training on our dataset achieves superior predictions on textureless regions (top row; blank ceiling), nature details (middle row; background leaves) and non-Lambertian surfaces (bottom row; TV screen). Images are from InStereo2k \cite{bao2020instereo2k}, Flickr1024 \cite{wang2019flickr1024}, and Booster \cite{zamaramirez2024booster}.}
    \label{fig:qual}

\end{figure*}

\parbf{Ratio of scene types} In Tab. \ref{tab:data_distribution}, we experiment on the distribution of scene types using the same experimental setup of a 5k stereo pair budget and training RAFT-Stereo for 75k steps. We find that a mixed dataset of all scene types is the most effective for stereo generalization. We also observe that indoor floating object scenes are the best individual scene type on all benchmarks except ETH3D. This result verifies the effectiveness of combining realistic scenes with floating objects over random flying objects without a background. The dense floating object dataset likely performs better on ETH3D because it has lower average disparity values and more closely matches the benchmark's distribution.

\subsection{Data Generation Cost Optimization}

Generating data with the default Infinigen configurations is computationally expensive. We aim to reduce the two largest costs: CPU time for generating the scene, and GPU time for rendering images from the generated scene.

\begin{table}[ht]
\centering
\scriptsize
\begin{tabular}{lcccccc}
\toprule
Experiment & \multicolumn{2}{c}{Midd-2014} & Midd & ETH3D & \multicolumn{2}{c}{KITTI} \\
& (F) & (H) & 2021 & & 12 & 15 \\
\midrule
\multicolumn{7}{l}{\emph{Ray-tracing Samples}}\\
8192 & 8.78 & 6.09 & \textbf{9.61} & 3.69 & 9.09 & 10.69 \\
1024 & 9.06 & 6.90 & 10.54 & 4.17 & 7.86 & 8.60 \\
1024 + denoise & 9.19 & 6.60 & 10.28 & 3.92 & 4.05 & \textbf{5.11} \\
1024 + denoise$^{\dagger}$ (Ours) & \textbf{7.92} & \textbf{5.63} & 9.88 & \textbf{2.91} & \textbf{3.62} & 5.45 \\
\midrule
\multicolumn{7}{l}{\emph{Solver Realism}}\\
Full Solve & 9.90 & 6.15 & \textbf{9.76} & 3.19 & 4.23 & 6.29 \\
Reduced Solve & 9.19 & 6.60 & 10.28 & 3.92 & 4.05 & \textbf{5.11} \\
Reduced Solve $^{\dagger}$ (Ours) & \textbf{7.92} & \textbf{5.63} & 9.88 & \textbf{2.91} & \textbf{3.62} & 5.45 \\
\bottomrule
\end{tabular}

\caption{\textbf{Cost optimization study}. We use the same experimental setup as Tab. \ref{tab:parameter_analysis_quantitative} except for those labeled with $\dagger$, which denotes the generation \textit{compute-matched} setting. Since our optimizations yield a ~6× speedup, $\dagger$ corresponds to 30k samples instead of 5k. While these optimizations reduce performance, they lower dataset cost and lead to better performance in the fixed-compute regime.}
\label{tab:cost_opt}
\vspace{-6pt}
\end{table}
\parbf{Indoor arrangement quality} Infinigen's room solver uses a simulated annealing algorithm that iteratively adds, removes, and moves objects to optimize a set of realism constraints. This process is slow, taking on average 50.85 minutes per indoor scene on 4 Xeon Gold 5320 CPUs. Naively reducing solver step count results in significantly sparser scenes, since the solver has fewer steps to place objects. We therefore constrain the solver to greedily add objects, which prevents the solver from wasting steps on moving and removing objects. This approach allows for significantly faster generation of densely populated scenes, at the cost of less realistic placement. We decrease the number of solver steps from 550 to 60, which reduces generation times to an average of 13 minutes. We observe that reducing solver realism leads to performance drops on Middlebury2014-H, Middlebury2021, and ETH3D, but not KITTI (Tab. \ref{tab:cost_opt}). This result suggests that more regular and realistic background scenes may help with in-domain performance, but hurt out-of-domain generalization. In the fixed-compute setting, our low-cost settings achieve better or comparable performance.

\parbf{Ray-tracing quality} Reducing sample count will linearly decrease render time, but results in high amounts of render noise in images. We propose the use of Blender's OptiX denoising algorithm to mitigate render noise. The denoising process results in reduced render fidelity and occasional artifacts, but our experimentation demonstrates that it is effective at improving the trade-off between render speed and render quality. We set render samples to 1024, which reduces rendering time to just 27 seconds per frame.

We observe that, without denoising, higher render sample counts result in improved performance on all benchmarks except KITTI, suggesting that improved render quality generally improves performance (Tab. \ref{tab:cost_opt}). We hypothesize that the large amounts of render noise function as an aggressive augmentation that aids generalization to the noisier images in the KITTI benchmark, but hurts performance on domains which have cleaner captures. This is similar to findings in \cite{Mayer_2018}, which reports small performance on KITTI benefits from explicitly modeling interpolation artifacts. By applying denoising to low-sample renders, model performance generally improves across most benchmarks, demonstrating that render denoising is effective at mitigating performance losses caused by excessive render noise. 

\parbf{Scene generation amortization} To reduce CPU cost, we aim to render more frames from each scene without sacrificing diversity. Traditionally, stereo datasets render \textit{video} trajectories from scenes. However, video sequences contain less diversity due to high overlap between nearby frames. Because we focus on two-view stereo matching, we achieve maximum diversity when frames are not contiguous. In each indoor scene, we place 20 separate stereo rigs, maximizing the standard deviation of depth values. For each dense floating object scene, we randomize object positions and orientations, lighting, and camera rig baseline values 200 times per scene. This allows us to generate fewer scenes, reducing CPU cost with a minimal loss in diversity.

\begin{table*}[ht] 
\centering 
\begin{tabular}{lccccccc} 
\toprule 
Dataset & \multicolumn{2}{c}{Middlebury 2014} & Middlebury & ETH3D & \multicolumn{2}{c}{KITTI} & Booster (Q) \\ 
& (F) & (H) & 2021 & & 2012 & 2015 & \\ 
\midrule 
SceneFlow   & 10.96 & 6.20 & 8.44 & 23.12 & 9.45 & 15.74 & 18.17 \\ 
CREStereo   & 14.45 & 11.53 & 10.6 & 5.18 & 4.95 & 5.90 & 14.61 \\ 
TartanAir   & 12.56 & 7.27 & 14.47 & 4.35 & 3.98 & 5.33 & 18.14 \\ 
IRS         & 7.81 & 6.13 & 8.49 & 3.91 & 4.56 & 5.60 & 10.32 \\ 
FSD         & 5.80 & \underline{3.27} & 6.93 & \underline{2.13} & \underline{3.56} & \textbf{4.18} & \underline{7.51} \\ 
\rowcolor{lightorange}\dataname{} 
            & \textbf{5.10} & 3.76 & \textbf{6.72} & 2.50 & \textbf{3.30} & 4.54 & 9.09 \\ 
\rowcolor{lightorange}FSD+\dataname{} 
            & \underline{5.24} & \textbf{3.24} & \underline{6.88} & \textbf{2.08} & 3.59 & \underline{4.26} & \textbf{7.42} \\ 
\bottomrule 
\end{tabular} 
\caption{\textbf{Zero-shot performance by dataset}. DLNR trained only on \dataname{} outperforms the same model trained on widely used datasets. Combining our data with FSD leads to best overall performance, demonstrating that our data can augment existing datasets.} 
\label{tab:dataset_results} 
 
\end{table*}

\begin{figure}
    \centering
    \includegraphics[width=1\linewidth]{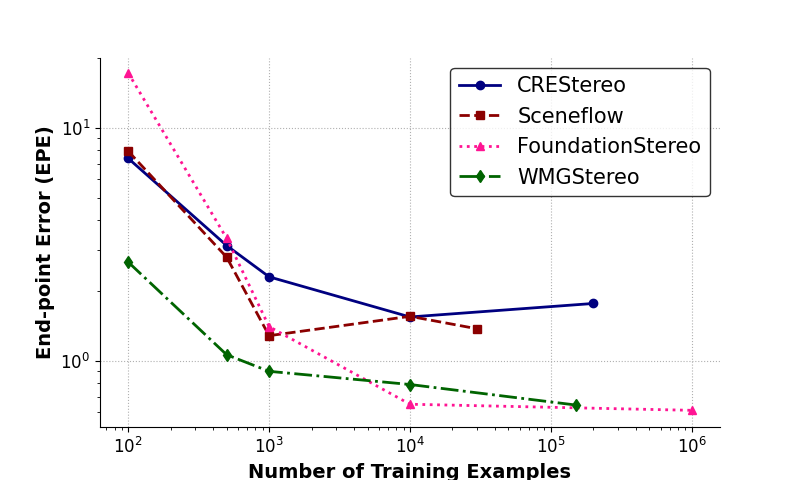}
    \caption{\textbf{Zero-shot performance by dataset size}. Our dataset is more \textit{sample-efficient} than SceneFlow and CREStereo, achieving better performance across a range of dataset sizes. Our dataset achieves competitive performance with FSD.}
    \label{fig:scaling}

\end{figure}

\section{\dataname{} Dataset}

We use the best settings in Tab. \ref{tab:parameter_analysis_quantitative} to construct \dataname{}, a stereo dataset of 163,396 stereo pairs. This dataset incorporates all the best parameter configurations found in Sec. \ref{sec:analysis}. We include three categories of rendered scenes, shown in Fig. \ref{fig:dataset_examples}: indoor floating objects (71,800 stereo pairs), dense floating objects (70,300 stereo pairs), and nature scenes (21,296 stereo pairs).

The total data generation cost was approximately 70.5 days on 4 CPUs for scene generation and 120 GPU days for rendering. Without any optimizations, it would have taken 275 days on 4 CPUs and 900 days on 1 GPU. In practice, we parallelize these computations over many CPUs and GPUs, drastically reducing wall-clock time to generate the dataset.

We train RAFT-Stereo \cite{lipson2021raft}, DLNR \cite{zhao2023high}, and Selective-IGEV \cite{wang2024selective} on our dataset for 200k steps from a random initialization, following the standard training procedures and hyperparameters used by the original authors. During training, we reweight the scene types on our dataset to sample each one equally, following results from Section \ref{sec:analysis}. During training, we mask the sky and untextured room exteriors. 

\subsection{Zero-shot Generalization}

Tab. \ref{tab:stereo_results} shows quantitative results for public models. Because most existing models are pre-trained only on SceneFlow, we retrain them on a mix of SceneFlow, TartanAir, CREStereo, and IRS to provide a more robust baseline (denoted -Mixed). DLNR-\dataname{} outperforms DLNR-Mixed by 28\% on Middlebury and by 25\% on the Booster Benchmark. RAFT-\dataname{} outperforms RAFT-Mixed by 11\% on KITTI 2012 and by 14\% on KITTI 2015, while Selective-IGEV-\dataname{} outperforms Selective-IGEV-Mixed by 20\% on Booster. Notably, RAFT-\dataname{} outperforms StereoAnywhere on Middlebury 2014, despite not leveraging any large monocular depth priors such as DepthAnythingV2 \cite{yang2024depthv2}.

\subsection{Qualitative Results}

In Fig. \ref{fig:qual}, we compare the qualitative results of a DLNR model trained on our dataset to the same model trained on other synthetic datasets. In the first row, we observe that our model accurately predicts difficult textureless regions. In the second row, our model is able to recover fine details from a natural scene, demonstrating that our model has not overfit to indoor domains. The third row shows that our model is robust to specular regions, likely due to \dataname{}'s photorealistic rendering and lighting.

\subsection{Dataset Comparison}

We also compare our dataset individually to existing datasets. For each of SceneFlow, CREStereo, TartanAir, and IRS, we train DLNR for 200k steps using default training settings. We report quantitative zero-shot results in Tab. \ref{tab:dataset_results}, using metrics as defined in Sec. \ref{sec:analysis}. The model trained on \dataname{} achieves better performance than classic datasets, with a 39\% improvement over IRS on Middlebury 2014 (H). Despite not explicitly modeling driving scenarios like FoundationStereo, our dataset achieves better performance on KITTI-12. Training on a combination of FSD and \dataname{} results in better performance than either dataset alone on Middlebury (H), ETH3D, and Booster, showing our data is useful when mixed with other datasets.

To demonstrate that our dataset has not overfit to the benchmarks in the study, we evaluate on held out benchmarks not used in parameter tuning in Tab. \ref{tab:held_out}. We use the Middlebury 2014 train set, excluding the 10 validation scenes, and a fixed, random subsample of 200 pairs from DrivingStereo \cite{Yang_2019_CVPR}. Training on our data results in a 27\% improvement over FSD in 3px error on DrivingStereo.

\begin{table}[ht]
\centering
\small
\begin{tabular}{lcccc}
\toprule
Dataset & \multicolumn{2}{c}{Midd-Train} & \multicolumn{2}{c}{DrivingStereo} \\
        & 2px & EPE & 3px & EPE \\
\midrule
DLNR-Sceneflow   & 24.23 & 18.49 & 18.65 & 4.79 \\
DLNR-CREStereo   & 22.07 & 9.30  & 8.21  & 1.62 \\
DLNR-IRS         & 16.36 & 7.21  & 3.89  & 1.06 \\
DLNR-TartanAir   & 23.15 & 13.01 & 2.40 & 0.86 \\
DLNR-Mixed       & 16.25 & 5.94  & 2.74  & 0.90 \\
DLNR-FSD         & 14.23 & 5.43 & 2.59  & 0.89 \\
\rowcolor{lightorange}DLNR-\dataname{}        & \textbf{12.61} & \textbf{4.54} & \textbf{1.89} & \textbf{0.82} \\
\bottomrule
\end{tabular}
\caption{\textbf{Zero-shot performance by dataset on held-out benchmarks} Training on \dataname{} results in better performance on benchmarks not used in parameter adjustment, demonstrating our findings generalize to unseen scenarios.}
\label{tab:held_out}

\end{table}

We also demonstrate in Fig. \ref{fig:scaling} that our data is sample-efficient. Given a fixed dataset size, we train DLNR for 200k steps on \dataname{}, SceneFlow, and CREStereo and evaluate their zero-shot performance on end-point error (EPE) on Middlebury 2014 (H). We find that just 500 training examples from \dataname{} achieve a lower EPE than 100k CREStereo samples. Moreover, our procedural ``recipe" continues to scale well as the number of training examples increases to over 100k samples, while SceneFlow plateaus at around 1000 samples and CREStereo at 10000 samples. Our dataset converges to a similar error as FSD, and demonstrates competitive sample efficiency.

\section{Conclusion}
We contribute a thorough study on \textit{what matters in stereo dataset generation}. By carefully selecting the best design choices, we generate a new synthetic dataset and demonstrate that training solely on this dataset achieves stronger zero-shot performance than training on many widely used datasets combined. We believe that our study and open-source codebase will be continually useful for stereo matching research, especially in co-developing training data with new architectures.

\section{Acknowledgments}

This work was partially supported by the National Science Foundation and an Amazon Research Award.

{
    \small
    \bibliographystyle{ieeenat_fullname}
    \bibliography{main}
}

\clearpage

\section*{Appendix}
\label{sec:appendix}
\section{Additional Analysis}
\textbf{Parameter Study} In Tab. \ref{tab:appendix_quant}, we provide an additional standard metric, end-point-error (EPE), for the same parameter study experiments in the main paper. For most benchmarks and experiments, the best model by px accuracy is also the best by EPE, demonstrating the robustness of our results.

\section{Additional Generation Details}

\subsection{Indoor Floating Objects Scene Type}

 For each scene, we first generate the indoor scene using our modified Infinigen Indoors settings. Given this indoor scene, we place the random objects in the scene, respecting collisions with existing scene objects. We then place the 20 camera rigs using the default Infinigen Indoors placement algorithm, which attempts to maximize the standard deviation of depth values. Finally, we apply augmentations as described in Section \ref{sec:analysis}. We randomize the materials of all floating and background objects, replacing an object's default material with a randomly chosen material with probability 0.5. To account for the large variance in object sizes, we normalize the floating object sizes.

\subsection{Dense Floating Objects Scene Type}
For each scene, we first spawn 200 objects into a scene at the origin, and use Blender's key-frame animation system to assign each object different locations, orientation, etc. at every single keyframe. Objects were placed at a distance sampled from Uniform[5,50] m away from the camera and the camera baselines were sampled from Uniform[0.1, 0.4] m. Similar to indoor scenes, we place point lights at random distances and normalize floating object size. We rescale object size based on distance to prevent far-away objects from being completely occluded by nearby objects and apply random warping to objects. We also perform additional lighting augmentations, randomizing the sky background's strength and color. 

\subsection{Nature Scene Type}
We generate nature scenes with the Infinigen Nature system, using default configurations unless otherwise specified in this section and \ref{sec:analysis}. By default, the Infinigen system provides configurations to render videos from a camera trajectory placed within a scene. However, these video sequences have high amounts of overlap between consecutive frames, so we render videos from camera trajectories at 6 frames per second to reduce similarity between consecutive frames. We render 50 frames from each video and sample the baseline of the stereo rig from Uniform[0.075, 0.5].

\subsection{Ground Truth}
For all frames, we provide camera intrinsics and extrinsics, left and right depth maps, and occlusion maps. We convert depth maps to disparity using the camera intrinsics and extrinsics. We compute occlusion maps by checking left-to-right and right-to-left re-projection consistency, with a 1 percent difference threshold. We also provide object segmentation maps, where each pixel has a value associated with an object's index value. Because procedural objects are never reused, we provide object names associated with each index, which enables the creation of flexible semantic segmentation maps. Similarly, materials are also procedural and never reused, so we provide material segmentation maps with material names. All frames are rendered at 1280 x 720 resolution.

\subsection{Dataset post-processing} Certain types of undesirable scenes are hard to prevent during dataset generation, so we provide tools to remove them as a post-processing step. We use a simple mean of pixel intensities to remove frames that are extremely dark as a result of lighting augmentation. We filter frames depth value lower than 12.5 cm to eliminate scenarios where the camera partially clips into an objects. We optionally apply a more aggressive distance filter of 5 m for the nature spit to filter out extremely high disparity samples, and release both the filtered and unfiltered nature split.

\subsection{Atmospheric effects}
We disable rain, dust, and snow particles from generation because they produce depth values that are almost invisible in rendered images. Similarly, we disable the coral reef scene type because of insufficiently visible background objects.

\clearpage %
\subsection{List of Assets Used}

\noindent\begin{adjustbox}{max width=\columnwidth,
                           max totalheight=\dimexpr\textheight-5\baselineskip\relax,
                           valign=t}
\normalsize
\setlength{\tabcolsep}{3pt}
\renewcommand{\arraystretch}{0.9}
\begin{tabular}{@{}ll@{}}
ArmChair & Balloon \\
BarChair & BasketBase \\
Bathtub & BathroomSink \\
Bed & BedFrame \\
BeverageFridge & Blanket \\
BookColumn & Book \\
BookStack & Bottle \\
Bowl & Can \\
CeilingClassicLamp & CellShelf \\
Chair & Chopsticks \\
Cup & CurvedStaircase \\
DeskLamp & Dishwasher \\
FloorLamp & FoodBag \\
FoodBox & Fork \\
GlassPanelDoor & Hardware \\
Jar & KitchenCabinet \\
KitchenIsland & KitchenSpace \\
Knife & Lamp \\
LargePlantContainer & LargeShelf \\
Lid & LiteDoor \\
LouverDoor & Mattress \\
Microwave & Monitor \\
NatureShelfTrinkets & OfficeChair \\
Oven & Pallet \\
Pants & PanelDoor \\
Pillow & PlantContainer \\
Plate & Pot \\
Rug & Shirt \\
SimpleBookcase & SimpleDesk \\
Sink & SingleCabinet \\
Sofa & SpiralStaircase \\
Spoon & StraightStaircase \\
TableCocktail & TableDining \\
Tap & Towel \\
TriangleShelf & TV \\
TVStand & Vase \\
WallArt & Wineglass \\
UShapedStaircase & \\
\end{tabular}
\end{adjustbox}

\begin{table*}
    \centering
    \scriptsize
    \setlength{\tabcolsep}{3pt}
    \renewcommand{\arraystretch}{1.2}
\begin{tabular}{l l *{14}{c}}
\toprule
\multirow{3}{*}{Experiment} & \multirow{3}{*}{Method} &
\multicolumn{4}{c}{Middlebury 2014} &
\multicolumn{2}{c}{Middlebury 2021} &
\multicolumn{2}{c}{ETH3D} &
\multicolumn{4}{c}{KITTI} &
\multicolumn{2}{c}{Booster (Q)} \\
\cmidrule(lr){3-6}\cmidrule(lr){7-8}\cmidrule(lr){9-10}\cmidrule(lr){11-14}\cmidrule(lr){15-16}
 &  & \multicolumn{2}{c}{H} & \multicolumn{2}{c}{F} & \multicolumn{2}{c}{--} &
   \multicolumn{2}{c}{--} & \multicolumn{2}{c}{2012} & \multicolumn{2}{c}{2015} &
   \multicolumn{2}{c}{--} \\
\cmidrule(lr){3-4}\cmidrule(lr){5-6}\cmidrule(lr){7-8}\cmidrule(lr){9-10}\cmidrule(lr){11-12}\cmidrule(lr){13-14}\cmidrule(lr){15-16}
 &  & px & EPE & px & EPE & px & EPE & px & EPE & px & EPE & px & EPE & px & EPE \\
\midrule
\multirow{3}{*}{Floating Object Density}
& No floating Objects           & 12.52 & 3.91 & 18.44 & 12.13 & 16.31 & 3.29 & 4.47 & 0.36 & 4.42 & 0.93 & 6.19 & 1.20 & 16.40 & 4.14 \\
& 0 to 10 floating objects      &  7.78 & 2.19 & 11.42 &  5.28 & 11.30 & 1.66 & \textbf{3.62} & 0.31 & 4.44 & 0.91 & 6.09 & 1.15 & 12.21 & 2.64 \\
& \textbf{10 to 30 floating objects} &  \textbf{6.60} & \textbf{1.77} & \textbf{9.19} & \textbf{4.37} & \textbf{10.28} & \textbf{1.55} & 3.92 & \textbf{0.30} & \textbf{4.05} & \textbf{0.86} & \textbf{5.11} & \textbf{1.05} & \textbf{10.60} & \textbf{2.20} \\
\midrule
\multirow{2}{*}{Background Objects}
& No background objects         &  8.35 & 2.15 & 10.42 &  5.58 & 12.01 & 1.68 & 4.39 & 0.35 & 4.20 & 0.92 & 6.28 & 1.21 & 12.72 & 3.34 \\
& \textbf{With background objects} & \textbf{6.60} & \textbf{1.77} & \textbf{9.19} & \textbf{4.37} & \textbf{10.28} & \textbf{1.55} & \textbf{3.92} & \textbf{0.30} & \textbf{4.05} & \textbf{0.86} & \textbf{5.11} & \textbf{1.05} & \textbf{10.60} & \textbf{2.20} \\
\midrule
\multirow{3}{*}{Floating Object Type}
& Floating chairs               &  \textbf{5.29} & 1.50 &  9.24 &  5.84 &  9.81 & 1.41 & 3.64 & 0.29 & 4.90 & 1.01 & 7.02 & 1.22 & 11.22 & 2.71 \\
& Floating shelves              &  7.13 & 1.89 & 10.28 &  5.05 & 10.24 & 1.43 & 3.51 & 0.31 & 4.32 & 0.94 & 6.06 & 1.16 & 11.63 & 2.31 \\
& Floating bushes              &  6.04 & \textbf{1.48} & 9.21 & 3.83 & \textbf{9.54} & \textbf{1.40} & \textbf{3.13} & \textbf{0.28} & \textbf{3.86} & 0.86 & 5.42 & 1.11 & 12.19 & 2.72  \\
& \textbf{All generators used}  & 6.60 & 1.77 & \textbf{9.19} & \textbf{4.37} & 10.28 & 1.55 & 3.92 & 0.30 & 4.05 & \textbf{0.86} & \textbf{5.11} & \textbf{1.05} & \textbf{10.60} & \textbf{2.20} \\
\midrule
\multirow{4}{*}{Object Material}
& No materials                  &  9.02 & 2.32 & 11.42 &  6.37 & 10.65 & 1.56 & 3.48 & 0.32 & 4.34 & 0.93 & 6.07 & 1.18 & 14.07 & 3.67 \\
& One diffuse material          &  7.21 & 1.82 & 10.09 &  5.28 &  \textbf{9.65} & \textbf{1.43} & \textbf{2.77} & \textbf{0.29} & \textbf{3.76} & \textbf{0.84} & 5.41 & 1.09 & 12.73 & 3.62 \\
& Only metal and glass          &  8.37 & 2.06 & 11.28 &  6.47 & 11.85 & 1.73 & 4.95 & 0.34 & 4.06 & 0.86 & \textbf{4.97} & \textbf{1.04} & \textbf{9.80} & \textbf{1.60} \\
& \textbf{All materials used}   & \textbf{6.60} & \textbf{1.77} & \textbf{9.19} & \textbf{4.37} & 10.28 & 1.55 & 3.92 & 0.30 & 4.05 & 0.86 & 5.11 & 1.05 & 10.60 & 2.20 \\
\midrule
\multirow{3}{*}{Stereo Baseline}
& Uniform[0.04, 0.1]            &  9.60 & 4.03 & 32.47 & 33.02 & 22.18 & 14.70 & \textbf{2.89} & \textbf{0.27} & 5.13 & 1.05 & 6.64 & 1.22 & 17.03 & 5.22 \\
& Uniform[0.2, 0.3]             &  7.01 & 2.17 &  9.75 &  5.39 & 10.50 &  1.72 & 14.05 & 6.00 & \textbf{3.94} & \textbf{0.86} & 5.37 & 1.10 & \textbf{8.96} & \textbf{1.76} \\
& \textbf{Uniform[0.04, 0.4]}   & \textbf{6.60} & \textbf{1.77} & \textbf{9.19} & \textbf{4.37} & \textbf{10.28} & \textbf{1.55} & 3.92 & 0.30 & 4.05 & \textbf{0.86} & \textbf{5.11} & \textbf{1.05} & 10.60 & 2.20 \\
\midrule
\multirow{2}{*}{Lighting}
& Realistic Lighting            &  6.91 & \textbf{1.74} &  9.62 & \textbf{4.36} & \textbf{10.06} & \textbf{1.46} & \textbf{3.81} & 0.31 & \textbf{3.95} & 0.88 & 5.45 & 1.10 & \textbf{10.45} & \textbf{2.09} \\
& \textbf{Augmented Lighting}   & \textbf{6.60} & 1.77 & \textbf{9.19} & 4.37 & 10.28 & 1.55 & 3.92 & \textbf{0.30} & 4.05 & \textbf{0.86} & \textbf{5.11} & \textbf{1.05} & 10.60 & 2.20 \\
\bottomrule
\end{tabular}
    \caption{\textbf{Expanded metrics for parameter study}. We compute EPE on all valid pixels.}
    \label{tab:appendix_quant}
\end{table*}

\begin{figure*}
    \centering
    \includegraphics[width=1\linewidth]{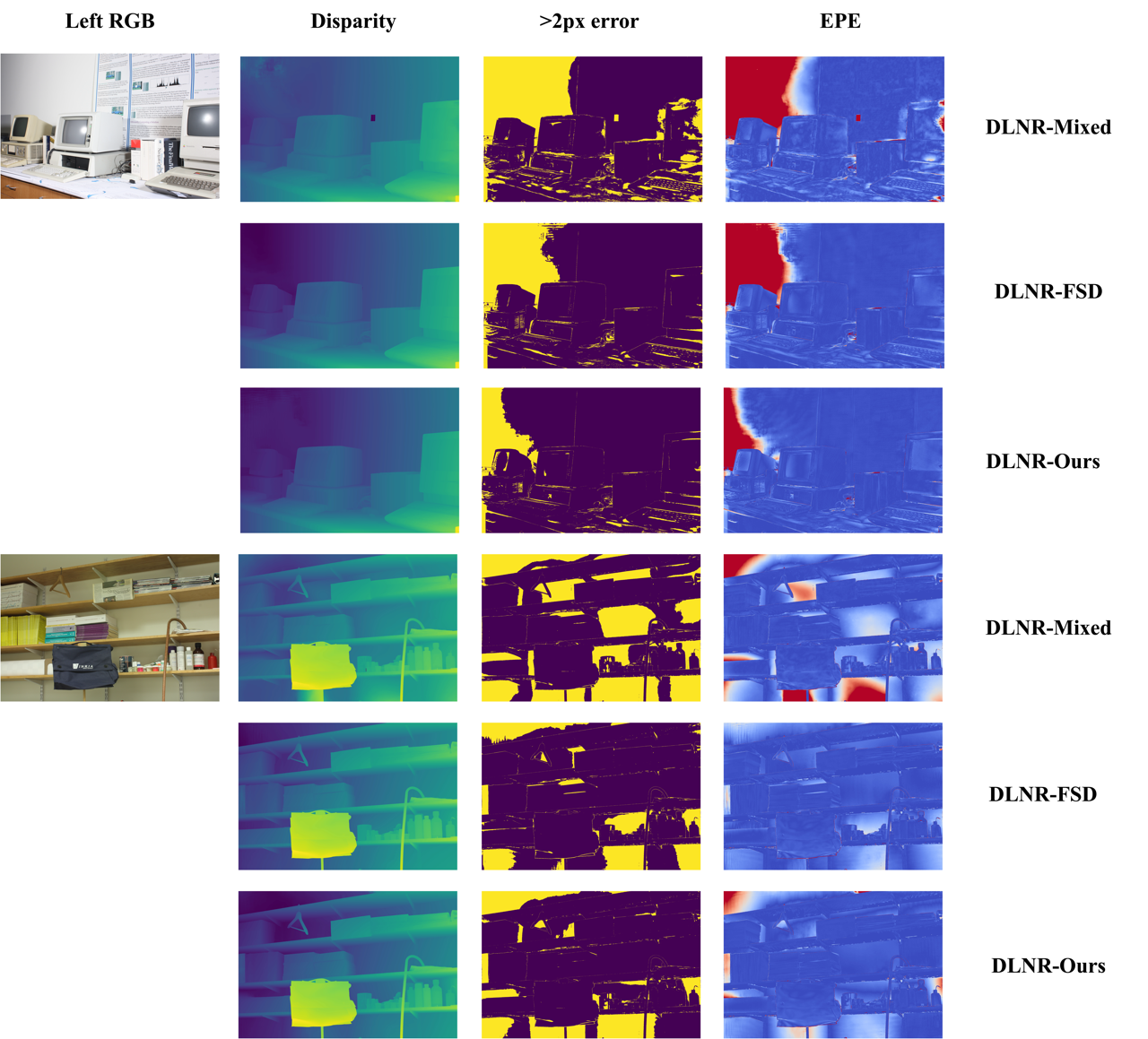}
    \caption{Qualitative results comparing DLNR-\dataname{}, DLNR-FSD, and DLNR-Mixed on the 2014 Middlebury Eval set. Training on our dataset results in more accurate capture of walls and textureless regions than DLNR-Mixed and comparable performance with DLNR-FSD.}
\end{figure*}

\begin{figure*}
    \centering
    \includegraphics[width=1\linewidth]{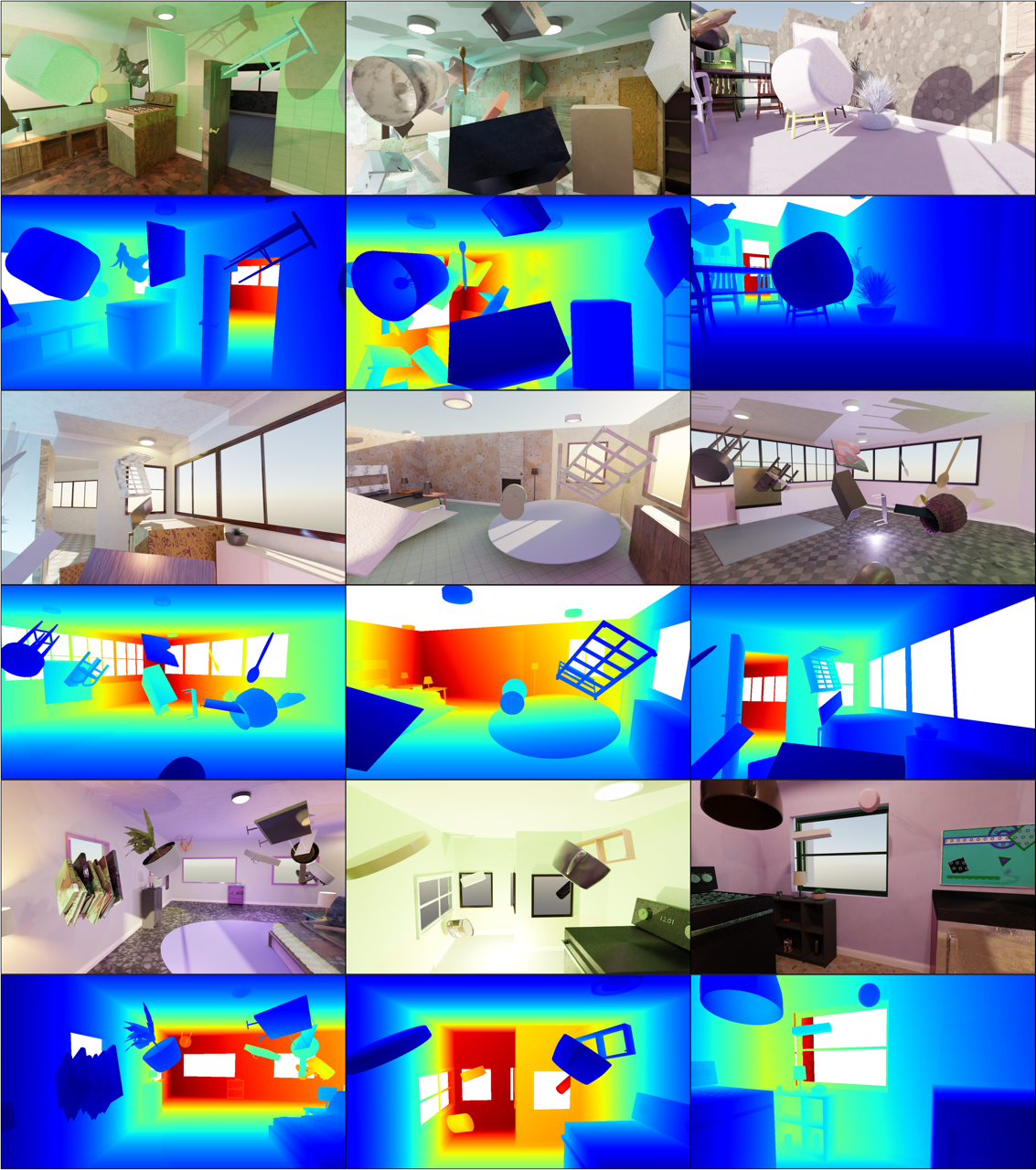}
    \caption{Non-cherrypicked, random samples from our floating indoor dataset type. We remove the glass portion of the window mesh entirely, instead of replacing the material, in order to maintain good lighting, as specified in Section \ref{sec:analysis}. This modification is a configurable option, and new data can be generated with the window meshes in place.}
\end{figure*}

\begin{figure*}
    \centering
    \includegraphics[width=1\linewidth]{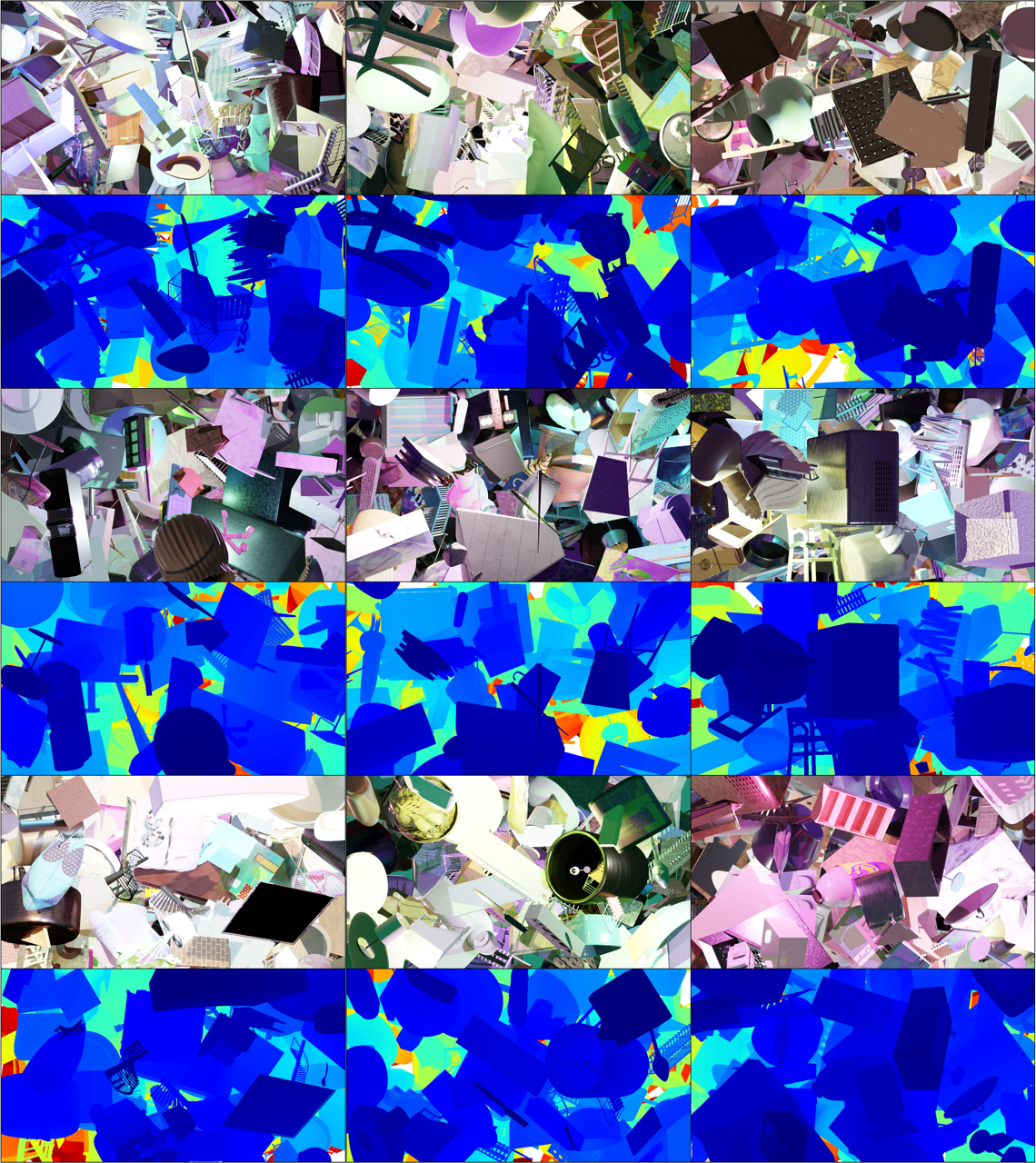}
    \caption{Non-cherrypicked, random samples from our dense floating dataset type.}
\end{figure*}

\begin{figure*}
    \centering
    \includegraphics[width=1\linewidth]{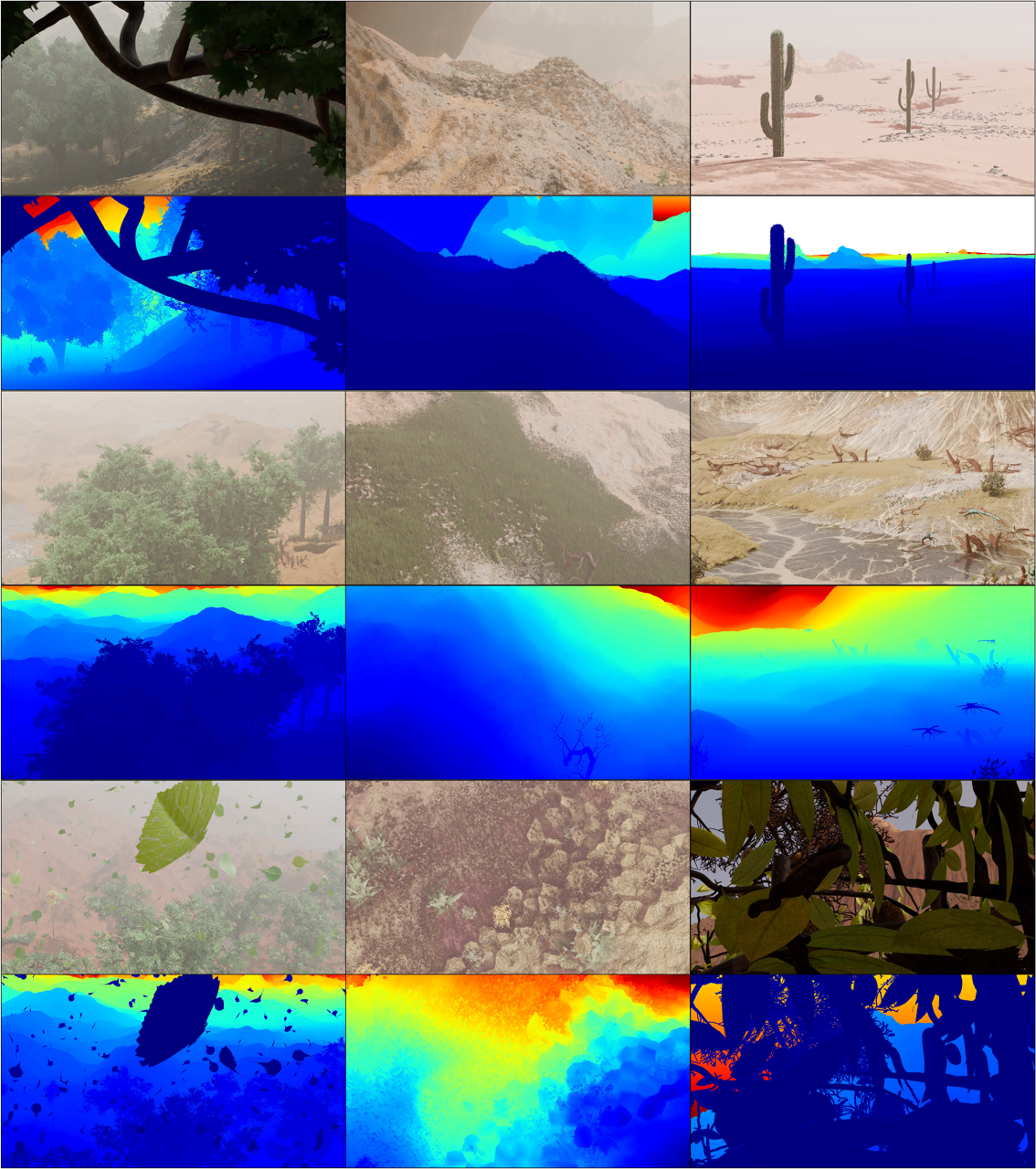}
    \caption{Non-cherrypicked, random samples from our nature dataset type.}

\end{figure*}

\end{document}

%% file: preamble.tex
\usepackage[dvipsnames]{xcolor}